\documentclass{article} % For LaTeX2e
\usepackage{iclr2024_conference,times}
\usepackage{graphicx} 
\usepackage{times}
\usepackage{latexsym}
\usepackage{booktabs}
\usepackage{enumitem}
\usepackage{multirow}
\usepackage{amssymb}
\usepackage{array}
\usepackage{amsmath}
\usepackage{multicol}
\usepackage{multirow}
\usepackage{wrapfig}
\usepackage{caption}
\usepackage{amsmath,amsfonts,bm}

\def\eqref#1{equation~\ref{#1}}
\def\1{\bm{1}}

\DeclareMathAlphabet{\mathsfit}{\encodingdefault}{\sfdefault}{m}{sl}
\SetMathAlphabet{\mathsfit}{bold}{\encodingdefault}{\sfdefault}{bx}{n}

\usepackage{hyperref}
\usepackage{url}

\usepackage{float}
\usepackage[capitalise]{cleveref}
\usepackage[group-minimum-digits={3}]{siunitx}
\usepackage{colortbl}
\definecolor{lg}{gray}{0.89}

\newcommand{\mytt}[1]{{\tt{#1}}}

\usepackage{microtype}
\newcommand*{\affmark}[1][*]{\textsuperscript{#1}}
\newcommand*{\affaddr}[1]{#1}

\usepackage{listings}
\definecolor{rebut}{HTML}{000000}
\newcommand{\rebut}[1]{\textcolor{rebut}{#1}}

\title{Zero and Few-shot Semantic Parsing with \\ Ambiguous Inputs}

\author{Elias Stengel-Eskin\affmark[1] 
        \quad 
        Kyle Rawlins\affmark[2]
        \quad 
        Benjamin Van Durme\affmark[2] \\ 
        \affaddr{\affmark[1]UNC Chapel Hill} \quad 
        \affaddr{\affmark[2]Johns Hopkins University} \\
        }

\iclrfinalcopy % Uncomment for camera-ready version, but NOT for submission.
\begin{document}

\maketitle

\begin{abstract}
Despite the frequent challenges posed by ambiguity when representing meaning via natural language, it is often ignored or deliberately removed in tasks mapping language to formally-designed representations, which generally assume a one-to-one mapping between linguistic and formal representations. 
We attempt to address this shortcoming by introducing \textsc{AmP}, a framework, dataset, and challenge for translating ambiguous natural language to formal representations like logic and code. 
We define templates and generate data for five well-documented linguistic ambiguities.
Using \textsc{AmP}, we investigate how several few-shot text-to-code systems handle ambiguity, introducing three new metrics.
We find that large pre-trained models perform poorly at capturing the distribution of possible meanings without deliberate instruction.
However, models are able to capture the distribution well when ambiguity is attested in their inputs. 
These results motivate a call for including ambiguity explicitly in datasets and promote considering the distribution of possible outputs when evaluating systems. \footnote{Data and code: \url{https://github.com/esteng/ambiguous_parsing}. 
Contact: \mytt{esteng@cs.unc.edu}
}
\end{abstract}

\section{Introduction}
\vspace{-1em}
Formalizing the meaning of natural language into a symbolic representation has been attempted across a variety of domains, from philosophy and linguistics \citep{wittgenstein.l.1921, montague.r.1970} to artificial intelligence \citep{winograd.t.1972, zelle.j.1996}. 
Attempts at formalization have often faced a shared challenge: many natural language statements have multiple possible meanings, i.e. they are ambiguous. 
Past work \citep[e.g.][]{zipf.g.1949, piantadosi.s.2012} has argued that this is a natural feature of a communication system, resulting from competing pressures on speakers and listeners.
Specifically, \citeauthor{piantadosi.s.2012} contend that ambiguity allows speakers to minimize their efforts.
Rather than exactly specify their intended meaning (resulting in a long and expensive message), speakers can send shorter, cheaper messages and rely on listeners to resolve any ambiguities.
However, this resolution in turn relies on \emph{commonsense knowledge} and \emph{conversational context}: most speakers of English would infer from the utterance, \textit{``I ate spaghetti with a fork''} that someone used a fork as a utensil, but commonsense knowledge would preclude this parse of \textit{``I ate spaghetti with meatballs''}. 
Similarly, conversational context can provide clues to help us choose between interpretations. 

Language can be used not only to communicate with other people, but also to interact with AI agents. 
One common method for interaction is semantic parsing, whereby natural language is translated into a formal and symbolic representation of its meaning (e.g. code, logic, graphs, etc.). 
However, human tools for ambiguity resolution may be unavailable to these non-human translation systems: models typically lack human-like commonsense knowledge and are missing conversational context. 
\rebut{This could lead to miscommunications between humans and models.
Since semantic parsing systems are used to perform real-world actions (e.g. modifying a calendar, sending emails, controlling physical robots, etc.) ambiguity-based miscommunication in parsing could have real-world consequences.}
% Models of language lack the embodied world experience which underlies much of our commonsense knowledge.
% Furthermore, we often interact with models in unnatural ways and without providing a full conversational context. 

\begin{figure*}
    \centering
    \includegraphics[width=\textwidth]{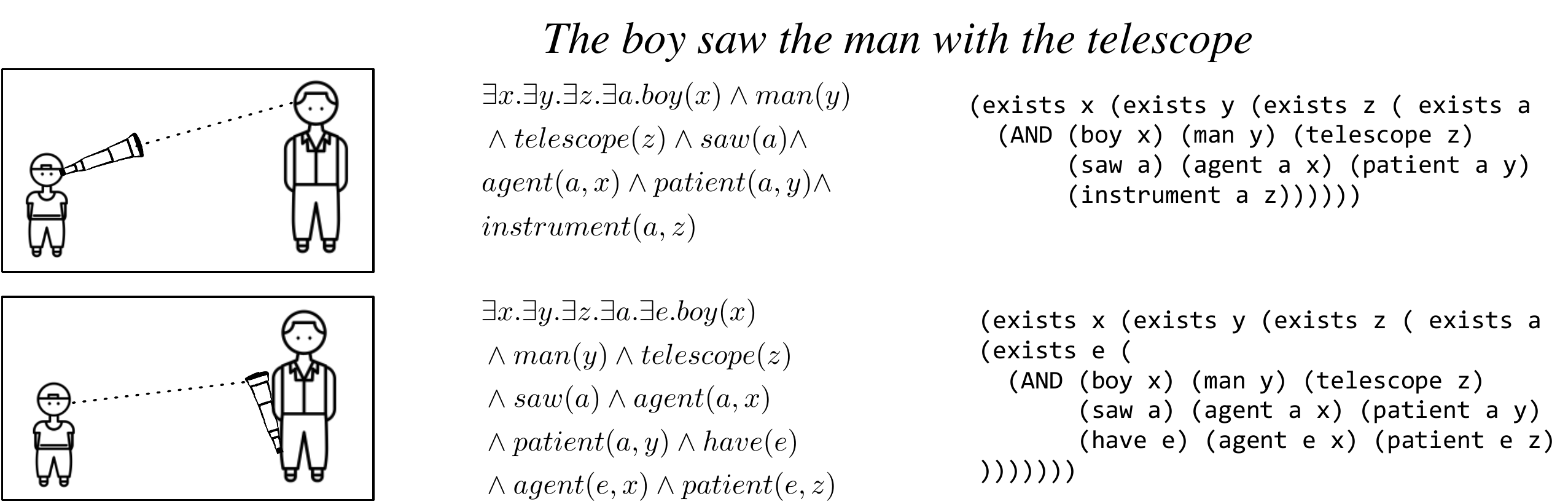}
    \vspace{-2em}
    \caption{An example of prepositional phrase (PP) attachment ambiguity. 
    The statement is compatible with two possible interpretations, represented visually, in first-order logic, and in Lisp format.}
    \vspace{-2em}
    \label{fig:fig1}
\end{figure*}

Ideally, given an ambiguous input, we would like our parsing models to capture a \emph{distribution} over interpretations with some uncertainty across plausible items in the distribution. 
This would allow robust handling of ambiguous utterances -- for example by enabling smart follow-up interactions \citep{stengel-eskin.e.2023didyoumean} -- getting us closer to the goal of using language as a general-purpose API for interaction.
Given that semantic parsing systems are typically based on language models -- which represent distributions over strings -- \rebut{combined with a search procedure (e.g. beam search)} it could be that models already capture ambiguity.
However, this hypothesis is hard to test given current semantic parsing datasets, which typically commit to a single interpretation for each utterance. 
To this end, we introduce an extensible framework and dataset for investigating ambiguity in semantic parsing. 
Our framework consists of templates covering five well-documented types of natural language ambiguity: prepositional-phrase attachment, scope and inverse scope, pronominal coreference, and conjunctions. 
For each type, our templates can generate large numbers of ambiguous and unambiguous utterances.
Each ambiguous utterance is paired with two possible interpretations, or logical forms (LFs); see \cref{fig:fig1} for an example.
% For example, in \cref{fig:fig1}, the two interpretations for, ``The boy saw the man with the telescope'' are visualized and their LFs are given.
LFs can be represented as first-order-logic (FOL) formulae or as programs in Lisp. 
We use our framework to create a benchmark dataset we call \textsc{AmP} (\textbf{Am}biguous \textbf{P}arsing). 
Unlike past efforts which have grounded ambiguous utterances in answers to questions \citep{stengel-eskin.e.2023}, language inferences \citep{liu.a.2023}, videos \citep{berzak.y.2015}, or images \citep{mehrabi.n.2022}, we focus our dataset on semantic parses.
This choice follows from several motivating factors. 
For one, semantic parsing has a long tradition of use in interactive systems, including in robotics \citep{kate.r.2005, tellex.s.2011, artzi.y.2013, tellex.s.2020}, question-answering \citep{zelle.j.1996, berant.j.2013, yu.t.2018}, and digital assistants \citep{semanticmachines2020, damonte.m.2019}. 
Ensuring that these systems capture ambiguity and that their confidence reflects appropriate uncertainty about the user's intent is crucial, as misunderstandings could have negative real-world consequences \citep{stengel-eskin.e.2023didyoumean}.
Furthermore, semantic parsing not only allows people to access computation, but also provides a way for models to use external tools: for example, simple forms of semantic parsing have been employed to augment large language models (LLMs) \citep{parisi.a.2022, schick.t.2023, mialon.g.2023}.  
Finally, long-form text-to-code shares many challenges with parsing.

Using our generated \textsc{AmP} data, we introduce a pair of challenging tasks designed for LLMs using in-context learning (ICL). 
In ICL, rather than explicitly training models to predict LFs, we provide models with instructive examples in a prompt, which is prepended to the test input.
This parsing setting has become increasingly popular in semantic parsing \citep{shin.r.2021, shin.r.2022, roy.s.2022}. 
Our tasks aim to quantify how well existing models capture ambiguity and to provide a framework for improving their ability to predict multiple meanings. 
We develop 3 metrics to measure models' performance on ambiguity in two settings: \textbf{zero-shot} and \textbf{mixed prompt}. 

In the \textbf{zero-shot} setting, we provide models with the ``ingredients'' to produce both possible derivations of a given ambiguity type, but we provide no examples of that ambiguity type; see \cref{append:prompt} for an example. 
In this unique compositional generalization challenge, the model must combine structures into novel derivations and \emph{also} recognize that the structures can be selected and combined in two ways to produce different derivations. 
We also annotate a subset of our data with crowdsourced judgements, comparing these to our models' predictions.
Models struggle to predict parses correctly in this setting.
When they do compose parses correctly, although models and people tend to choose similar interpretations, models generally fail to predict both possible parses. 

In the \textbf{mixed prompt} setting, we examine how model distributions and outputs change when varying the number of examples for each interpretation in the prompt. 
For each ambiguity type, we construct ``mixed prompts'' consisting of conflicting examples. 
Some examples shown to the model pair utterances of an ambiguity type with one kind of LF, and others pair the same kinds of inputs with the alternative LF. 
This setting is motivated by a case in which ambiguity might lead to conflicting annotations in a training dataset; when examples are retrieved from that data to construct a prompt for a test example, the resulting prompt will also contain conflicting parses. 
Here, our metrics measure to what extent a model represents the distribution in its input given conflicting evidence.
Some models perform remarkably well here, aligning with the prompt distribution across ambiguity types. 
To our knowledge, this is the first study of in-context learning with conflicting evidence. 
\vspace{-0.5em}

\begin{table*}
    {\footnotesize
    \begin{tabular}{p{0.093\linewidth} | p{0.185\linewidth}p{0.3\linewidth}p{0.3\linewidth}}
    \hline
    Type & Ex. Input & $LF_0$ & $LF_1$ \\
    \hline
    Prep. phrase attachment (PP) & \emph{The man saw the boy with the telescope} & $\exists x . \exists y . \exists z . \exists a . \exists e . man(x) \land boy(y) \land saw(a) \land agent(a, x) \land patient(a, y) \land telescope(z) \land have(e) \land agent(e, y) \land patient(e, z)$ \emph{Interpretation}: the man saw the boy, who was holding a telescope.  &  $\exists x . \exists y . \exists z . \exists a . man(x) \land boy(y) \land  telescope(z) \land saw(a) \land agent(a, x) \land patient(a, y) \land instrument(a, z)$  \emph{Interpretation}: the man used a telescope to see the boy.\\
    \hline
    Quantifier scope (Scope) & \emph{every cow saw a dog} & $\exists x . \forall y . \exists a . cow(y) \land dog(x)  \land saw(a)  \land agent(a, y)  \land patient(a, x)$ \emph{Interpretation}: there is \emph{exactly} one dog. & $\forall x . \exists y . \exists a . cow(x)  \land dog(y)  \land saw(a)  \land agent(a, x)  \land patient(a, y)$ \emph{Interpretation}: there may be more than one dog. \\
    \hline
    Reversed, or inverse scope (revscope) &  \emph{a cow saw every dog} &  $\exists x . \forall y . \exists a . cow(x) \land dog(y) \land saw(a) \land agent(a, x) \land patient(a, y)$  \emph{Interpretation}: there is \emph{exactly} one cow. & $\forall x . \exists y . \exists a . cow(y) \land dog(x) \land saw(a) \land agent(a, y) \land patient(a, x)$ \emph{Interpretation}: there may be more than one cow.\\
    \hline
    Pronoun coreference (bound) & \textit{Mary saw the woman and she smiled} & $\exists x . \exists a . \exists e . woman(x) \land saw(a) \land agent(a, Mary) \land patient(a, x) \land smiled(e) \land agent(e, Mary)$ \emph{Interpretation}: Mary smiled.
       &   $\exists x . \exists a . \exists e . woman(x) \land saw(a) \land agent(a, Mary) \land patient(a, x) \land smiled(e) \land agent(e, x)$ \emph{Interpretation}: the woman smiled. \\
    \hline
    Conjunction (conj.) & \emph{the man drank and ate or swam} & $\exists x . \exists a . \exists e . \exists i . man(x) \land ( ( drank(a) \land agent(a, x) \land ate(e) \land agent(e, x) ) \lor ( swam(i) \land agent(i, x) ) )$ \emph{Interpretation}: the man either drank and ate or he swam. & 
    $\exists x . \exists a . \exists e . \exists i . man(x) \land ( drank(a) \land agent(a, x) \land ( ( ate(e) \land agent(e, x) ) \lor ( swam(i) \land agent(i, x) ) ) )$ \emph{Interpretation}: the man drank, and he either ate or swam.  \\
    \hline
    \end{tabular}
    }
    \vspace{-1em}
    \caption{Ambiguity types considered with example inputs and LFs. See \cref{append:lex} for more description, including the lexical items used.}
    \label{tab:ambiguities}
    \vspace{-1.75em}
\end{table*}

% For certain ambiguity types, this can be done by varying the lexical items -- for example, \emph{the man saw the boy with the toy} is not ambiguous. 
% We create unambiguous statements which will be used in \cref{sec:zero_shot}, such as \emph{the man saw the boy} or \emph{the girl with the telescope}}. 

\vspace{-0.75em}
\section{Methods}
\vspace{-0.75em}
\paragraph{Data} 
We introduce a dataset of ambiguous parses, where natural language examples are parsed into first-order logic (FOL). 
Further details on the construction of our logical forms (LFs) can be found in \cref{append:data}.
We can canonicalize our LFs, so that logically equivalent formulae with varying syntax are treated as identical:
we transform LFs into binary trees, where nodes are ordered alphabetically, and we anonymize variables. 
% This means that the following formulae for \emph{the man walked the dog} are treated as identical: $\exists x . \exists y . \exists e . man(x) \land dog(y) \land walk(e) \land agent(e, x) \land patient(e, y)$, $\exists x . \exists y . \exists e . dog(x) \land man(y) \land walk(e) \land patient(e, x) \land agent(e, y)$,
% $\exists a . \exists w . \exists m . walk(a) \land man(w) \land dog(m) \land agent(a, w) \land patient(a, m)$.
Note that when prompting our model, we do use a standard variable set and order, where the variables $x,y,z$ are used for nouns, and $a, e, i$ are used for events. 
Our canonicalization process also allows us to render our LFs in different formats. 
In addition to a standard FOL format, we experiment with a Lisp format (cf. \cref{fig:fig1}).
For machine-readability, we always render logical connectives in plaintext, i.e. $\exists$ becomes \mytt{exists}, $\land$ becomes \mytt{AND}, etc..
We consider five types of syntactic and semantic ambiguities, given in \cref{tab:ambiguities}.\footnote{\rebut{We can also generate unambiguous examples, and can extend AMP to new ambiguity types/vocab items.}} 

\vspace{-0.75em}
\paragraph{Models}
Semantic parsing tasks are often framed as sequence transduction, where a model learns to translate text into LFs by training on paired data \citep{dong.l.2016, zhang.s.2019b}. 
It has become clear that neural models can capture distributions they are trained on; thus, if we were to train on ambiguous data, it would not be surprising if the model captured ambiguity, and vice-versa.
Rather than training models, we instead consider several models for in-context learning (ICL), focusing on large pre-trained autoregressive (AR) language models. 
We use the Codegen series of models \citep{nijkamp.n.2022} --  350 million (M), 2 billion (B), 6B, and 16B parameters -- which are based on the GPT-2 architecture \citep{radford.a.2019} and are pre-trained on large amounts of code and text.\footnote{Past work \citep{shin.r.2022} has shown that code pretraining improves over text pre-training on other ICL semantic parsing tasks.} 
% We consider four sizes of Codegen models: 350 million (M) parameters, 2 billion (B), 6B, and 16B. 
We also use LLMs pre-trained on text; here, we examine Llama-13B \citep{touvron.h.2023}, an open-source AR transformer.
To examine the impact of instruction tuning \citep{wei.j.2022b}, we consider Vicuna-13B \citep{vicuna2023}, which uses prompts distilled from ChatGPT to instruction-finetune Llama-13B. 
All models above 350M were run at \mytt{fp16} precision.
In the zero-shot setting, we also consider OpenAI's gpt-3.5-turbo.
While few details about the model are known, it is a large AR transformer model which has undergone both instruction tuning and fine-tuning from human feedback \citep{ouyang.l.2022}.
It is often the most performant model; however, the API does not provide access to logit scores, precluding analyses of uncertainty. 
As such, we only use it in our zero-shot experiments, where the metric is accuracy-based rather than uncertainty-based. 
For non-API models, we used constrained decoding \citep{shin.r.2021, shin.r.2022, roy.s.2022} to ensure the model only produces valid logical statements; see \cref{append:decoding} for details.

\vspace{-0.75em}
\paragraph{Computing probability under a forced decode} \label{sec:compute_prob} 
In our analyses, we would like to compare the probabilities the model assigns to $LF_0$ and $LF_1$.
While one could compare the product of probabilities under the model for each LF, we find that in practice, this results in very low scores for either LF. 
We instead use \citet{stengel-eskin.e.2023calibration}'s sequence confidence estimate to obtain $P_\theta(LF_0)$, renormalizing at the end:  
\begin{enumerate}[noitemsep,nolistsep,topsep=0pt,leftmargin=*]
\vspace{-0.5em}
    \item We obtain token-level probabilities under a forced decode of $LF_0$ and $LF_1$. 
    For each token $t_i$ in an LF with tokens $t_1 \ldots t_N$, we compute $P_\theta(t_i |x;t_{1:i-1})$, where $t_{1:i-1}$ is the \emph{gold} token prefix and $x$ is the input prompt.
    \item We take $\hat{p} = \min_{i=1}^{N} P_\theta(t_i | x;t_{1:i-1})$, the minimum probability across all tokens.\footnote{We also experimented with averaging, which resulted in similar results. We chose min to be consistent with \citet{stengel-eskin.e.2023calibration}.}
    \item We normalize the probabilities: $P_\theta(LF_0) = \frac{\hat{p}_{LF_0}}{\hat{p}_{LF_0} + \hat{p}_{LF_1}}$ and set $P_\theta(LF_1) = 1-P_\theta(LF_0)$
\end{enumerate}

\vspace{-1em}
\subsection{Metrics} 
\vspace{-0.5em}

\paragraph{Zero-shot metrics}
In the zero-shot setting, we aim to measure the degree to which the model captures both possible interpretations of an ambiguous statement. 
Intuitively, when given the ``ingredients'' to make both interpretations, a model that robustly captures ambiguity should allocate some probability to both. 
We measure this by computing the proportion of elements for which the model has both interpretations in its top-$k$ predictions. 
Note that we remain agnostic here to the exact probability of each interpretation; we aim instead to quantify whether it predicts both interpretations at all.
Note also that as we increase $k$, this metric becomes less stringent. 
Let $T_k$ be the top $k$ most probable predictions from the model under some sampling method (e.g. beam search), and $\mathbb{I}$ be an indicator function. 
The zero-shot metric $ZM_k$ is given by \cref{eqn:zeroshot_dataset_metric}.
This metric counts how often both LFs are found in the top $k$ outputs \rebut{averaged across examples $i \in [1, N]$.}. 
It ranges from 0 to 100, \rebut{and higher is better, as it indicates more examples have both LFs in their top $k$ outputs. } 
\vspace{-0.75em}
\begin{align}
    &ZM_k = \frac{ \sum\limits_{i=1}^{N} \big( \mathbb{I}[LF_0 \in T_k] * \mathbb{I}[LF_1 \in T_k] \big)}{N}*100 \label{eqn:zeroshot_dataset_metric} 
\end{align}
\vspace{-0.5em}

\vspace{-0.5em}
\paragraph{Few-shot metrics}
In the few-shot setting, we are concerned about the level to which the model is capturing the distribution given in the prompt.
A core assumption here is that an ideal model would perfectly capture the uncertainty in the given distribution. 
We consider metrics at two levels of granularity to evaluate this behavior.
The first metric we consider measures model performance at the level of the dataset. 
Intuitively, as we sweep across ratios $r\in R$, we expect the proportion of predicted LFs to match $r$.
For example, when $r = 0.10$ (meaning that $10\%$ of the prompt examples are $LF_0$ and $90\%$ are $LF_1$) we would expect the model to produce $LF_0$ in roughly $10\%$ of instances. 
Let $y_i$ be the predicted LF for input instance $x_i$. Then the fewshot dataset metric $FDM$ is given by \cref{eqn:fewshot_dataset_metric}. 
Intuitively, this measures the difference between the accuracy on each LF and the ratio of that LF; lower is better for $FDM$, which ranges from $1.0$ to $0.0$.
The second metric measures model performance at the level of individual datapoints. 
If the model is capturing the distribution in the prompt, then the probability assigned to $LF_0$ should roughly match $r$, e.g. if $r = 0.10$, the model should assign $P(LF_0) \approx 0.10$. 
The few-shot instance metric $FIM$ is given by \cref{eqn:fewshot_instance_metric}.
$FIM$ resembles a Brier score \citep{brier.g.1950} and measures the error between the predicted probability and the ratio; it also ranges from $1.0$ to $0.0$ and lower is better. 
\vspace{-0.75em}
\begin{align}    
    &FDM = \frac{1}{|R|} \sum\limits_{r\in R} \Big(   \big| \big( \frac{1}{N} \sum\limits_{i=1}^{N}\mathbb{I}[y_i = LF_0]\big) - r \big|  + 
    % &\phantom{FDM = }+ 
    \big| \big( \frac{1}{N} \sum\limits_{i=1}^{N} \mathbb{I}[y_i = LF_1]\big) - (1-r)   \big| \Big)  \label{eqn:fewshot_dataset_metric} \\
    &FIM = \frac{1}{|R|} \sum\limits_{r\in R} \big(\frac{1}{N} \sum\limits_{i=1}^{N}  
        (P_\theta(y_i = LF_0) - r)^2 
    \big) \label{eqn:fewshot_instance_metric}
\end{align}

\vspace{-1em}
\section{Experiment 1: Zero-shot parsing} \label{sec:zero_shot} 
\vspace{-0.5em}
For each ambiguity type, we construct a prompt that provides the ingredients for deriving both LFs. 
The order of the component sentences is shuffled to avoid biasing the model towards one interpretation or the other. 
Crucially, the prompt contains no examples of the types of sentences being tested. 
% \cref{fig:zero_shot_prompt} shows an example for PP attachment.
For example, for PP attachment, the model is given an example of how to parse transitive verbs (\emph{``the boy saw the man''}), instruments (\emph{``the boy saw with the telescope''}), and possessives (\emph{``the boy with the telescope''}), each in isolation. 
To successfully generalize, the model has to overcome two challenges: first, it must compositionally generalize to compose the ingredients in the prompt into a valid derivation.
Secondly, it must recognize the ambiguity and reflect both derivations in its output. 
For each ambiguity type, we test 200 examples. 
Prompt examples are given in \cref{append:prompt}.

\vspace{-0.5em}
\subsection{Zero-shot results and analysis} \label{sec:zeroshot_res} 
\begin{figure}[ht]
    \centering
    \centering
    \vspace{-1em}
    \includegraphics[width=\textwidth]{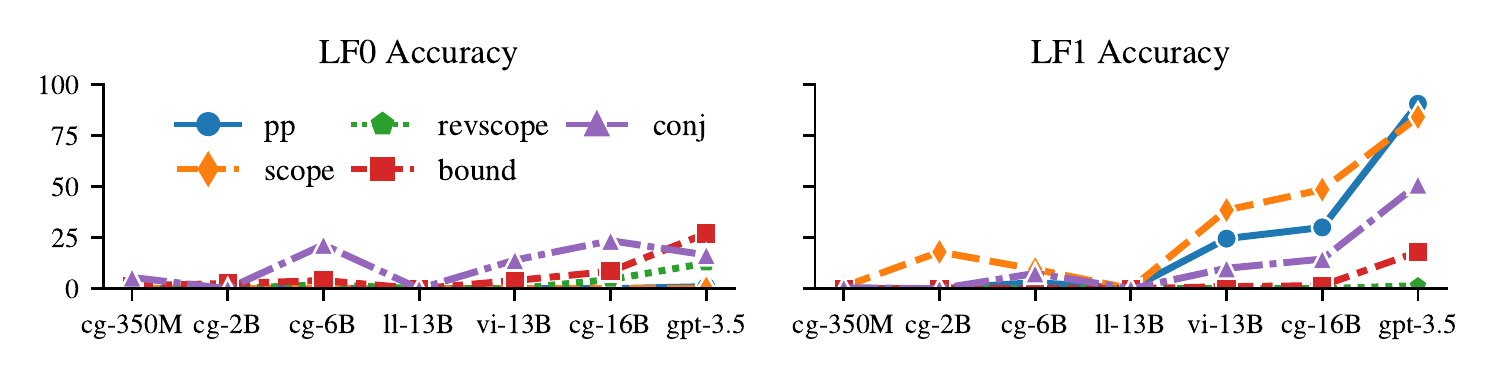}
    \vspace{-3em}
    \caption{Zero-shot exact-match accuracy on ambiguity types. Cg = Codegen, Ll = Llama, Vi = Vicuna. Models increasing in size from left to right. } 
    \vspace{-0.5em}
    \label{fig:zero_shot_acc} 
\end{figure}

\vspace{-0.5em}
In \cref{fig:zero_shot_acc}, smaller Codegen models (350M, 2B) struggle to predict either LF correctly.
On some ambiguity types, larger Codegen models (6B, 16B) predict one LF correctly.
\rebut{However, most models fail to ever predict both LFs correctly; exceptions to this are conjunction and coreference ambiguities, where we see some models correctly predicting $LF_0$ for some examples and $LF_1$ for others.}
GPT-3.5 does well at predicting $LF_1$ for PP and scope ambiguities, and predicts both LFs for coreference and conjunctions.
Interestingly, while Llama-13B is unable to correctly predict either LF for any of the ambiguities, Vicuna-13B (Llama's instruction-tuned variant) is comparable to Codegen-16B, suggesting that instruction tuning helps the model predict one LF correctly (though not to capture ambiguity).
Separately, we find that predicting FOL generally outperforms Lisp; for example, the Codegen-2B model on scope predicts $LF_1$ correctly $18\%$ of the time when using FOL and only $11\%$ when using Lisp; we report only FOL results moving forward. 

\begin{figure}[t]
    \begin{minipage}{0.43\textwidth}
        \centering
        \includegraphics[width=\textwidth]{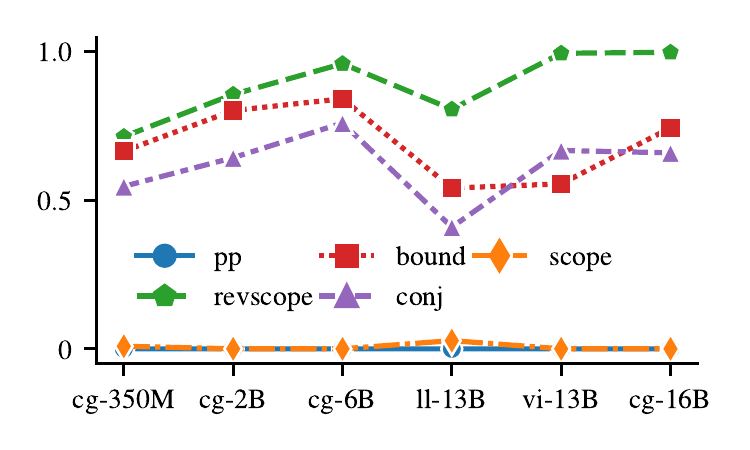}
        \vspace{-2em}
        \caption{$P(LF_0)$ per model.}
        \vspace{-0.5em}
        \label{fig:zero_shot_prob} 
    \end{minipage}
    \begin{minipage}{0.57\textwidth}
        {\footnotesize
        \begin{tabular}{cccccc}
            \hline
            Model & PP & Scope & Revscope & Bound & Conj. \\
            \hline
        cg-350M & 0.00 & 0.00 & 0.00 & 0.00 & 1.00 \\
        cg-2B & 0.00 & 0.00 & 0.00 & 0.50 & 0.00 \\
        cg-6B & 0.00 & 0.00 & 0.00 & 0.00 & 3.50 \\
        cg-16B & 0.00 & 0.00 & 0.00 & 3.50 & 15.00 \\
        ll-13B & 0.00 & 0.00 & 0.00 & 0.00 & 0.00 \\
        vi-13B & 0.00 & 0.00 & 0.00 & 4.00 & 9.50 \\
        gpt-3.5 & 0.00 & 0.00 & 0.00 & 0.00 & 0.00 \\
        \hline
        \end{tabular}
        }
        % \vspace{-1em}
        \captionof{table}{$ZM_5$ for all models (cg = Codegen). Models generally fail to predict both LFs.}
        \label{tab:zeroshot_metrics}
        \vspace{-1em}
    \end{minipage}
    \vspace{-0.75em}
\end{figure}

These results are further underscored in \cref{tab:zeroshot_metrics} showing the $ZM_5$ values for all models. 
We see that all models tested perform very poorly on this metric, with most models scoring $0.0$ in most settings. 
Qualitatively, we find that the model's top 5 outputs tend to include variations of the same LF. 
This finding aligns with the probability results seen in \cref{fig:zero_shot_prob}, where models tend to assign extreme probabilities to LFs on $3$ out of $5$ ambiguity types.
Notable exceptions here are conjunction and bound pronoun types, where in \cref{fig:zero_shot_prob} models assign closer to 0.5 probability to each parse; we see this also reflected in \cref{fig:zero_shot_acc} and \cref{tab:zeroshot_metrics}, where models predict both $LF_0$ and $LF_1$ correctly some of the time. 
As a whole, these results underscore the difficulty of the compositional task we have proposed; while some models are able to obtain high accuracy on one LF in isolation (GPT-3.5 predicts $LF_1$ for PP attachment and scope almost perfectly) no model is able to consistently predict both interpretations. 

\begin{figure*}[t]
    \centering
    % \vspace{-0.5em}
    \includegraphics[width=\textwidth]{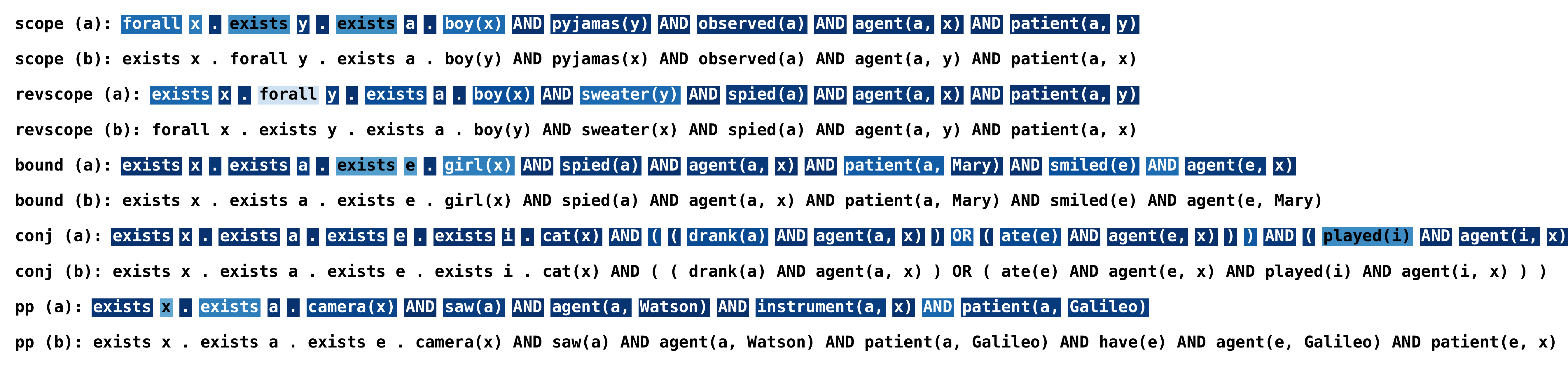}
    \vspace{-2em}
    \caption{Zero-shot per-token probability (darker is more probable) for each ambiguity type. Alternative parse given below each predicted parse. Token probability sometimes reflects divergences between the parses.}
    \vspace{-1em}
    \label{fig:heatmap}
\end{figure*}

We can also ask whether token-level confidence reflects the ambiguity in the space of possible parses.
Examining task-oriented semantic parsing models, \citet{stengel-eskin.e.2023calibration} find that many models (including Codegen) are relatively well-calibrated at the token level, meaning their confidence aligns with their average accuracy. 
Taking token-level probabilities as confidence scores, we follow their analysis and ask whether models are well-calibrated w.r.t. alternative parses. 
Specifically, we compare the model's confidence on tokens at the points where the predicted and alternative parse diverge. 
This is visualized for each ambiguity type in \cref{fig:heatmap}. 
Here, we take the first correctly-predicted LF (either $LF_0$ or $LF_1$) from Codegen-16B, predicted via beam search with grammar-constrained decoding (not forced decoding).
We overlay the confidence onto the token as the background color (darker is more confident). 
Below each predicted parse, we give the alternative parse.
% We see that for certain ambiguity types, the low-confidence tokens do correspond to points where the programs diverge.
For scope and inverse scope, the model assigns low confidence to the quantifier tokens at the start of the formula, which are in the reverse order in the alternative parse. 
Similarly, the tokens involving the quantified variables have lower confidence. 
We also see low confidence around the area of divergence for conjunction. 
However, pronominal coreference and PP attachment lack such interpretable confidence changes. 
These results are promising: for some ambiguity types, the model's token-level confidence reflects the alternative parse. 

\vspace{-0.5em}
\subsection{Human validation} \label{sec:human}
\vspace{-0.5em}
\cref{fig:zero_shot_acc} indicates that models tend to produce one interpretation or the other -- when models have non-zero accuracy on one interpretation, they tend to have zero on the other. 
Psycholinguistic research suggests that people have preferred interpretations \citep{anderbois.s.2012, dwivedi.v.2013}; at the aggregate level, \cref{fig:zero_shot_prob} shows that models align with human preferences on scope ambiguities. 

\citet{anderbois.s.2012} and \citet{dwivedi.v.2013} also describe strong lexical effects in scope ambiguity, meaning that the choice of words in the example has an effect on the interpretation taken. 
In order to further examine how the models tested compare with these results, we annotate a subset of our validation examples with human interpretations and confidence scores. 
This allows us to compare model predictions to humans at an item-level in addition to an aggregate level.

Annotators were asked to choose between interpretations and provide a confidence score on a sliding scale, following the EASL protocol \citep{sakaguchi.k.2018}; the confidence score was then converted to a probability (cf. \cref{append:annotation} for details). 
Since annotators are unlikely to know FOL or Lisp, each LF is shown as a statement that clearly indicates the interpretation (as in \cref{tab:ambiguities}). 
For example, for a PP attachment example like, \textit{``the boy saw the man with the telescope''}, the verbalized interpretations are \textit{``the boy saw the man, who had/was wearing a telescope''} and \textit{``the boy saw the man and used a telescope to do so''}. 
For each ambiguity type except conjunctions, we randomly select 20 examples from our development splits.\footnote{Conjunction ambiguities were excluded due to difficulties in verbalizing their interpretations fluently.}
Each example is annotated by 3 annotators.

\begin{figure}[ht]
    % \vspace{-1.25em}
    \begin{minipage}{0.5\textwidth}
        \centering
        \includegraphics[width=\textwidth]{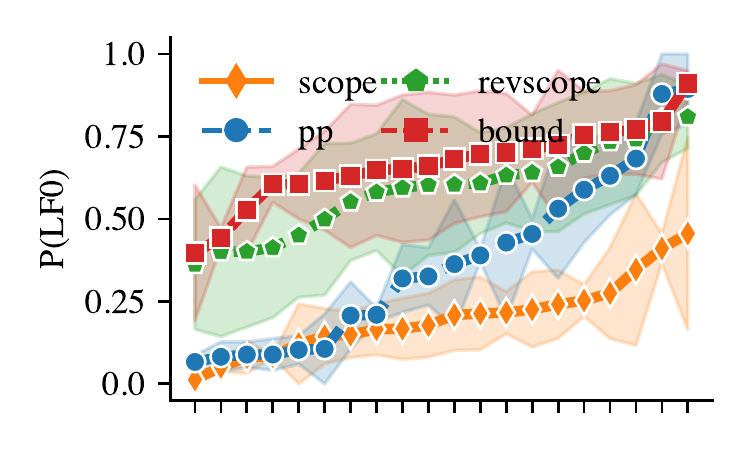}
    \end{minipage}
    \begin{minipage}{0.5\textwidth}
        \centering
        \includegraphics[width=\textwidth]{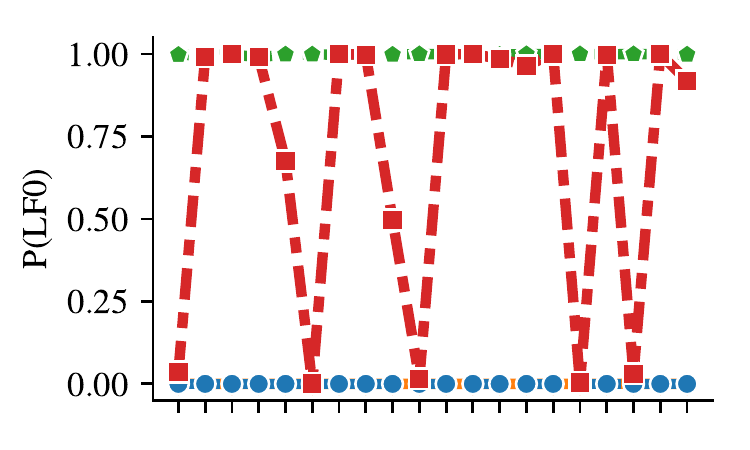}
    \end{minipage}
    \vspace{-2em}
    \caption{Per-example probabilities derived from humans (left) and cg-16B (right) on LFs. Examples are sorted by probability. Human probabilities vary according to vocabulary choice, but model probabilities generally do not. }
    \label{fig:hit_probs}
    \vspace{-1.5em}
\end{figure}

We find that annotators disagree almost as often as they agree: 38 examples have disagreement while 42 have all 3 annotators agreeing. 
This is a positive finding, indicating that our examples are highly ambiguous.
\cref{fig:hit_probs} (left) shows confidence scores (averaged across 3 annotators) for each item (sorted separately by mean confidence for each ambiguity type). 
There are broad preferences for all ambiguity types except PP attachment: bound and inverse scope tend to be matched to $LF_0$, and scope to $LF_1$. 
The latter aligns with some past findings \cite{anderbois.s.2012, caramazza.a.1977}.\footnote{Other work has found linear order to have a negligible effect and pointed to additional factors influencing interpretations \citep{kurtzman.h.1993, dwivedi.v.2013}.}
For PP attachment, some inputs are confidently parsed as $LF_0$ and others as $LF_1$. 
% Thus, lexical choices have an impact on the interpretation taken. 

In \cref{fig:hit_probs} (right), we contrast the human results with the output of the Codegen-16B model. 
While the probabilities generally match in direction to the human annotations (except for PP-attachment) we do not see the same kind of item-level sensitivity. 
For most PP, scope, and inverse scope examples, we see the model assigning all examples the same extreme probability. 
For bound pronouns, we see more variation, with the model switching between $LF_0$ and $LF_1$; however, the predictions are fairly extreme. 
These results suggest that the model is poorly-calibrated w.r.t. ambiguity at the item level.

\vspace{-0.5em}
\section{Experiment 2: Few-shot parsing} \label{sec:fewshot}
\vspace{-0.5em}
Ambiguities may lead to similar inputs being paired with different logical forms. 
Increasingly, it is common to retrieve examples from a training set to compose a prompt for ICL.
If that training set has ambiguity in it, it is likely that the retrieved prompt would contain conflicting examples, e.g. some examples pairing an utterance type with $LF_0$ and others pairing it with $LF_1$. 
In our few-shot experiments, we seek to fill this gap by investigating how model confidence and accuracy change at different prompt ratios. 
Crucially, by investigating mixed prompts with ambiguous inputs, we are ensuring that the disagreement in the prompt is not due to simple mistakes; one could imagine a mixed prompt arising from noisy data, where instances are mislabeled. 
In such cases, a strong enough model may even learn to ignore mislabeled data in the prompt.
\begin{wrapfigure}{r}{0.45\textwidth}
    \vspace{-1.5em}
    \includegraphics[width=0.45\textwidth]{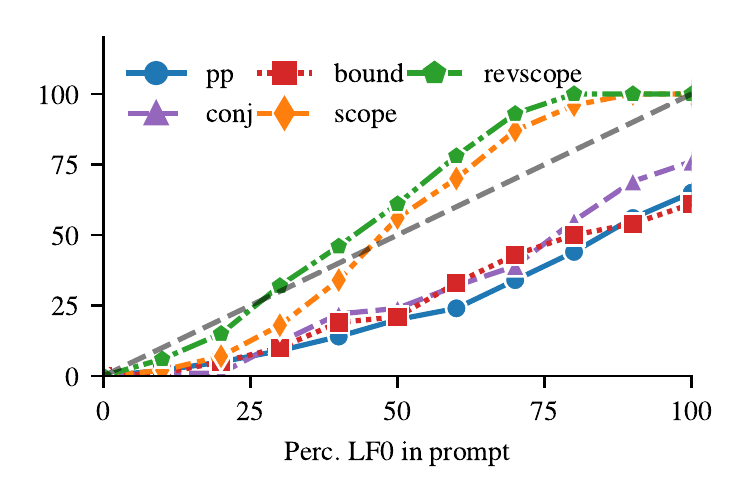}
    \vspace{-3em}
    \caption{Fewshot acc. increases according to the ratio of the LF in the prompt. $LF_1$ acc. (not pictured) decreases accordingly.}
    \vspace{-1.5em}
    \label{fig:few_shot_acc} 
\end{wrapfigure}
However, in the case of ambiguity, there are multiple \emph{legitimate} interpretations. 

For each ambiguity type, we construct prompts by pairing sentences of the same type with $LF_0$ in some cases, and $LF_1$ in others; we run 100 examples per type, per ratio. 
Each prompt contains 10 input-LF pairs, and a different prompt is constructed for each test sentence. 
\rebut{We vary the number of $LF_0$ sentences in the prompt from 0 to 10 in increments of 1 (e.g. $0-100\%$)} and shuffle the prompt sentences to ensure that there is no positional bias. 

\vspace{-0.5em}
\paragraph{Few-shot results and analysis} \label{sec:fewshot_res} 
\vspace{-0.5em}
\cref{fig:few_shot_acc} shows the accuracy of models on $LF_0$ as we increase the percentage of $LF_0$ in the prompt for the Codegen-2B model. 
We see that for scope and inverse scope, the accuracy tracks almost perfectly with the percentage. 
For other ambiguities, the accuracy is correlated with the percentage but never reaches $100\%$. 
\cref{tab:fewshot_metrics} shows the $FIM$ and $FDM$ scores for all models across all ambiguity types. 
Recall that $FDM$ measures how well the model's \emph{accuracy} aligns with the percentage of examples for each LF in the prompt: to obtain a lower $FDM$ score, the model needs to predict each LF at roughly the rate that it is seen in the prompt. 
Note that a model that fails to predict either LF correctly will have a high $FDM$ score.
For example, because Llama-13B fails to predict any of the LFs correctly, it has an $FDM$ score of $1.00$ for all ambiguities. 
For Codegen models, $FDM$ generally improves with model size on most ambiguity types. 
Overall, several models achieve fairly low $FDM$, especially on scope and inverse scope. 

We also see that all $FIM$ scores are relatively low. 
Because $FIM$ uses the gold LF to extract the sequence probability, even models with poor accuracy on both LFs can have a fairly low $FIM$. 
In other words, $FIM$ presents the model with a forced choice between two parses, rather than evaluating the model's most probable generations like $FDM$ and $ZM$ do. 
For example, Llama-13B never produces a correct LF under beam-search decoding with constraints, but when probabilities are extracted with a forced decode of the gold LFs, they align fairly well with those in the prompt, and it often attains lower $FIM$ scores than other models, including Vicuna. 
$FIM$ remains fairly constant with model size: Codegen-16B is tied with smaller models on 4/5 ambiguity types. 
Taken together, these results indicates that the models tested are surprisingly good at capturing the distribution in the prompt.
The low $FDM$ on some types indicates that overall, models like Codegen-16B and Vicuna-13B produce $LF_0$ roughly at the rate that it appears in the prompt.
However, we see that low $FIM$ does not imply low $FDM$, since low $FDM$ requires the model to be accurate. 

Interestingly, the models seem to override the zero-shot tendencies seen in \cref{fig:zero_shot_prob}, where scope and inverse scope were strongly associated with one LF over the other. 
With direct evidence on how to parse scope sentences in the prompt -- evidence we did not provide in the zero-shot setting -- the model produces the interpretations seen in the prompt, and is especially close to the prompt distribution for scope and inverse scope. 
These results are promising: given that models seem to capture mixed prompts well, it could be that ambiguity poses less of a challenge in settings where such mixed prompts can be constructed, i.e. settings with ambiguity in the training data. 

\begin{table*}[ht]
    \centering
    {\footnotesize
    \begin{tabular}{c|cc|cc|cc|cc|cc}
    \hline
    Model & \multicolumn{2}{c}{PP} & \multicolumn{2}{c}{Scope} & \multicolumn{2}{c}{Revscope} & \multicolumn{2}{c}{Bound} & \multicolumn{2}{c}{Conj.} \\
    \hline
    -- &  $FDM$ & $FIM$ &  $FDM$ & $FIM$ &  $FDM$ & $FIM$ &  $FDM$ & $FIM$ &  $FDM$ & $FIM$ \\
    \hline
    cg-350M & 0.76 & 0.08 & 0.19 & 0.05 & 0.36 & 0.06 & 0.62 & 0.16 & 0.60 & 0.06 \\
    cg-2B & 0.51 & 0.09 & 0.19 & 0.03 & 0.18 & 0.03 & 0.48 & 0.10 & 0.39 & 0.05 \\
    cg-6B & 0.45 & 0.08 & 0.21 & 0.05 & 0.16 & 0.06 & 0.43 & 0.10 & 0.41 & 0.06 \\
    cg-16B & 0.35 & 0.08 & 0.20 & 0.03 & 0.21 & 0.03 & 0.38 & 0.10 & 0.27 & 0.06 \\
    ll-13B & 1.00 & 0.06 & 1.00 & 0.04 & 1.00 & 0.04 & 1.00 & 0.06 & 1.00 & 0.05 \\
    vi-13B & 0.50 & 0.08 & 0.17 & 0.05 & 0.20 & 0.06 & 0.26 & 0.07 & 0.35 & 0.09 \\
    \hline
    \end{tabular}
    }
    \vspace{-0.5em}
    \caption{Few-shot metrics for all models (lower=better). $FDM$ (\cref{eqn:fewshot_dataset_metric}) measures the extent to which the model's accuracy across the whole dataset matches the percentage of that LF in the prompt. $FIM$ (\cref{eqn:fewshot_instance_metric}) measures how well the model's uncertainty captures the prompt's uncertainty.}
    \vspace{-1em}
    \label{tab:fewshot_metrics}
\end{table*}

\vspace{-0.5em}
\section{Related work}
\vspace{-0.5em}
Ambiguity has been a longstanding topic of interest in linguistics and psycholinguistics. 
Past work has argued that it is a feature arising naturally from the trade-off between competing objectives \citep{zipf.g.1949, schutze.h.1995, piantadosi.s.2012}.
However, providing a systematic account of ambiguity in linguistics lies beyond the scope of this section.

Some work in NLP has focused on modeling ambiguity in visual contexts, where questions and statements have been paired with images or videos depicting situations they refer to.
\citet{stengel-eskin.e.2023} introduce a dataset of linguistically ambiguous questions about images as well as a model for question disambiguation, and \citet{futeral.m.2022} examine ambiguous source sentences in machine translation, providing disambiguating images. 
More akin to our work, \citet{berzak.y.2015} introduce a corpus of syntactic, semantic, and pragmatic ambiguities with video interpretations. 
Follow-up work by \citet{mehrabi.n.2022} generates images of ambiguous statements.
We use many of the same ambiguities, but represent meaning with LFs instead of videos or images.
This is motivated in part by the relative ease of checking the correctness of a logical formula over, for example, an image.
Ambiguity has been studied in more general tasks such as question-answering \citep{min.s.2020}, natural language inference (NLI) \citep{liu.a.2023}, and coreference resolution \citep{yuan.y.2023}, where models have broadly been found lacking in their ability to resolve ambiguities. 
In parsing specifically, \citet{rasmussen.n.2020} introduce a $\lambda$-calculus dataset on 2,000 sentences of simple Wikipedia text, where roughly $50\%$ contain quantifier scope ambiguity.
We use synthetic data instead, giving us greater control and allowing us to examine more ambiguity types. 

\vspace{-0.5em}
\section{Discussion and Conclusion} \label{sec:discussion}
\vspace{-0.5em}
In addition to semantic parsing's many downstream applications, it has often been used to measure models' compositional generalization abilities through synthetic benchmarks like COGS \citep{kim.n.2020} and SCAN \citep{lake.b.2018}. 
The ability to generalize systematically and compositionally to unseen combinations is a core component of human intelligence \citep{fodor.j.1988}. 
Past efforts have generally assumed that there is a single correct LF for any given input, either implicitly \citep{lake.b.2018} or explicitly \citep[][Appendix H]{kim.n.2020}. 
This assumption is not borne out in natural language, where statements can be ambiguous and have multiple meanings.
It is also violated in many common applications of semantic parsing, such as text-to-code, where there are myriad ways of producing logically equivalent programs. 
Making this assumption reduces semantic parsing to syntactic parsing, \rebut{since there is a 1-to-1 mapping between syntax and meaning.}\footnote{For example, \citet{rudinger.r.2014} found that one of the key features separating dependency-based syntax and event-based semantics was the ability to handle PP attachment ambiguities.}
In future work, we hope to improve on the challenging and novel compositional task proposed in \cref{sec:zero_shot}, where all models struggle to capture both meanings.

\cref{sec:fewshot} offers a more hopeful takeaway: when ambiguity is present in the input, many models are able to capture the distribution of LFs.
Of course, for ambiguity to be attested in the prompt, it must exist in the data used to construct the prompt. \rebut{Furthermore, it needs to be attested in the evaluation data for us to test for it.}
However, in most current datasets, parses are not annotated redundantly or exhaustively, i.e. inputs are paired with a single output.
Given that annotators often disagree on ambiguous examples (cf. \cref{sec:human}) it is crucial to obtain multiple judgements, \rebut{at least on evaluation data}.\footnote{\rebut{For training data, we may be able to obtain \emph{diverse} judgements instead, with examples of a given ambiguity \emph{type} paired with different outputs, as was done in \cref{sec:fewshot}}}
This has recently become more common in other domains, such as NLI \citep{chen.t.2020, nie.y.2020, pavlick.e.2019}. 
Even when items are annotated redundantly, disagreement has often been discouraged or treated as noise. 
More recent work has begun to recognize that disagreement can arise for valid reasons \citep{pavlick.e.2019} including ambiguity \citep{bhattacharya.n.2019, stengel-eskin.e.2023, liu.a.2023}. 
\rebut{To improve the handling of ambiguity, we advocate for extending redundant, ambiguity-aware annotation protocols (with attention to disagreement) from single-label tasks (e.g. QA, NLI) to complex, sequential outputs like semantic parsing.}
\rebut{Improving both zero-shot generalization and data collection would help models capture the full range of utterance interpretations. 
This could lead to robust, interactive systems in which agents ask for confirmation or clarification on ambiguous examples \citep{stengel-eskin.e.2023didyoumean}, ultimately improving safety for critical systems.}

% Worth noting also is the link between ambiguity and underspecification.
Finally, ambiguous utterances are underspecified, i.e. they lack the requisite information to decide which interpretation is correct.
Some past work carries underspecification into the meaning representation: \citet{copestake.a.2005} introduces Minimum Recursion Semantics, which leaves noun-phrase bracketing (similar to conjunction ambiguities) and scope underspecified in the target representation, allowing multiple interpretations to be recovered. 
\citet{bos.j.2004} introduce a discourse representation that also maintain scope ambiguities for later resolution.
We have opted for a more fully-specified representation, placing the onus of resolution onto the parsing model rather than the representation.

\vspace{-0.75em}
\paragraph{Limitations} 
Firstly, we are limited by committing to a particular logical form. 
To test the parsing abilities of models under ambiguity, we are forced to choose a fixed meaning representation (MR) form, possibly including suboptimal abstractions and design choices. 
Motivated by \citet{wu.z.2023}, who find that arbitrary choices in an MR's construction can hamper compositional generalization, we have limited the number and difficulty of our choices, and mitigated their effect by offering two output formats.
It is also important to point out some key differences between the models we test and the relevant psycholinguistic literature. 
Broadly, experiments indicate that people maintain multiple interpretations during \emph{online} processing, later settling on one interpretation \citep{lackner.j.1972, rayner.k.1983, filik.r.2004}. 
Our models do not do online processing, and receive input as text, not audio. 
\rebut{Similarly, we do not provide a conversational context, which people generally use to resolve ambiguity.}
Thus, our results should not be taken to reflect how people process language. 

Methodologically, we are limited by our use of a fixed set of English-only ambiguities. 
We hope to add more languages, \rebut{lexical items,} and ambiguity types via our extensible data framework.
\rebut{Many optimizations have could be made for ambiguity, e.g. 
decoding strategies that emphasize diversity (temperature sampling, sequentially decoding outputs, etc.) might result in better results for the $ZM$ metric, since models lack output diversity. 
In our experiments we attempted to mimic how semantic parsing is commonly done in practice, without optimizing for ambiguity specifically.}

\vspace{-0.75em}
\paragraph{Conclusion}
By introducing a new benchmark for parsing under ambiguity, we are able to examine how modern semantic parsers handle cases where utterances can have multiple meanings. 
To this end, we introduced three new metrics for measuring the extent to which models capture the \emph{distribution} of meanings. 
While we find that models struggle to compose symbols without explicit guidance, we also find that they are sensitive the ambiguity when given mixed prompts, suggesting that having ambiguity in the training data may be a sufficient condition for capturing it in the output. 
This motivates our call for capturing ambiguity during annotation. 

\section{Ethics Statement}
By creating sentences templatically, we ensure that \textsc{AmP} does not contain any harmful texts. 
This protects our human annotators from exposure to such inputs; we also compensate the annotators at a level substantially above US federal minimum wage, and in-line with living wage estimates. 
Like most NLP research, our work has the potential to contribute to dual-use.
However, we believe that overall, making models robust to ambiguity will contribute to safer and more reliable technology, and has limited potential for negative applications.

\section{Reproducibility Statement}
In order to further reproducibility, we release our dataset and our code. 
This includes the code for \textsc{AmP}, which can be extended to generate new ambiguities, as well as the code for running all experiments. 
We do make use of a closed-source model (ChatGPT), which can hinder reproducibility; to hedge against this, most of our results are based on open-source models that are widely available. 
We also include our prompts in the appendix. 

% \section*{Acknowledgements}

\bibliography{calibration}

\begin{thebibliography}{63}
\providecommand{\natexlab}[1]{#1}
\providecommand{\url}[1]{\texttt{#1}}
\expandafter\ifx\csname urlstyle\endcsname\relax
  \providecommand{\doi}[1]{doi: #1}\else
  \providecommand{\doi}{doi: \begingroup \urlstyle{rm}\Url}\fi

\bibitem[Ander{B}ois et~al.(2012)Ander{B}ois, Brasoveanu, and
  Henderson]{anderbois.s.2012}
Scott Ander{B}ois, Adrian Brasoveanu, and Robert Henderson.
\newblock The pragmatics of quantifier scope: A corpus study.
\newblock In \emph{Proceedings of Sinn und Bedeutung}, volume~16, pp.\  15--28,
  2012.

\bibitem[Artzi \& Zettlemoyer(2013)Artzi and Zettlemoyer]{artzi.y.2013}
Yoav Artzi and Luke Zettlemoyer.
\newblock Weakly supervised learning of semantic parsers for mapping
  instructions to actions.
\newblock \emph{Transactions of the Association for Computational Linguistics},
  1:\penalty0 49--62, 2013.

\bibitem[Artzi et~al.(2015)Artzi, Lee, and Zettlemoyer]{artzi.y.2015}
Yoav Artzi, Kenton Lee, and Luke Zettlemoyer.
\newblock Broad-coverage ccg semantic parsing with amr.
\newblock In \emph{Proceedings of the 2015 Conference on Empirical Methods in
  Natural Language Processing}, pp.\  1699--1710, 2015.

\bibitem[Berant et~al.(2013)Berant, Chou, Frostig, and Liang]{berant.j.2013}
Jonathan Berant, Andrew Chou, Roy Frostig, and Percy Liang.
\newblock Semantic parsing on {F}reebase from question-answer pairs.
\newblock In \emph{Proceedings of the 2013 conference on empirical methods in
  natural language processing}, pp.\  1533--1544, 2013.

\bibitem[Berzak et~al.(2015)Berzak, Barbu, Harari, Katz, and
  Ullman]{berzak.y.2015}
Yevgeni Berzak, Andrei Barbu, Daniel Harari, Boris Katz, and Shimon Ullman.
\newblock Do you see what {I} mean? {V}isual resolution of linguistic
  ambiguities.
\newblock In \emph{Proceedings of the 2015 Conference on Empirical Methods in
  Natural Language Processing}, pp.\  1477--1487, 2015.

\bibitem[Bhattacharya et~al.(2019)Bhattacharya, Li, and
  Gurari]{bhattacharya.n.2019}
Nilavra Bhattacharya, Qing Li, and Danna Gurari.
\newblock Why does a visual question have different answers?
\newblock In \emph{Proceedings of the IEEE/CVF International Conference on
  Computer Vision}, pp.\  4271--4280, 2019.

\bibitem[Bos(2004)]{bos.j.2004}
Johan Bos.
\newblock Computational semantics in discourse: Underspecification, resolution,
  and inference.
\newblock \emph{Journal of Logic, Language and Information}, 13:\penalty0
  139--157, 2004.

\bibitem[Brier et~al.(1950)]{brier.g.1950}
Glenn~W Brier et~al.
\newblock Verification of forecasts expressed in terms of probability.
\newblock \emph{Monthly weather review}, 78\penalty0 (1):\penalty0 1--3, 1950.

\bibitem[Caramazza et~al.(1977)Caramazza, Grober, Garvey, and
  Yates]{caramazza.a.1977}
Alfonso Caramazza, Ellen Grober, Catherine Garvey, and Jack Yates.
\newblock Comprehension of anaphoric pronouns.
\newblock \emph{Journal of verbal learning and verbal behavior}, 16\penalty0
  (5):\penalty0 601--609, 1977.

\bibitem[Chen et~al.(2020)Chen, Jiang, Poliak, Sakaguchi, and
  Van~Durme]{chen.t.2020}
Tongfei Chen, Zhengping Jiang, Adam Poliak, Keisuke Sakaguchi, and Benjamin
  Van~Durme.
\newblock Uncertain natural language inference.
\newblock In \emph{Proceedings of the 58th Annual Meeting of the Association
  for Computational Linguistics}, pp.\  8772--8779, Online, July 2020.
  Association for Computational Linguistics.
\newblock \doi{10.18653/v1/2020.acl-main.774}.
\newblock URL \url{https://aclanthology.org/2020.acl-main.774}.

\bibitem[Chiang et~al.(2023)Chiang, Li, Lin, Sheng, Wu, Zhang, Zheng, Zhuang,
  Zhuang, Gonzalez, Stoica, and Xing]{vicuna2023}
Wei-Lin Chiang, Zhuohan Li, Zi~Lin, Ying Sheng, Zhanghao Wu, Hao Zhang, Lianmin
  Zheng, Siyuan Zhuang, Yonghao Zhuang, Joseph~E. Gonzalez, Ion Stoica, and
  Eric~P. Xing.
\newblock Vicuna: An open-source chatbot impressing gpt-4 with 90\%* chatgpt
  quality, March 2023.
\newblock URL \url{https://lmsys.org/blog/2023-03-30-vicuna/}.

\bibitem[Copestake et~al.(2005)Copestake, Flickinger, Pollard, and
  Sag]{copestake.a.2005}
Ann Copestake, Dan Flickinger, Carl Pollard, and Ivan~A Sag.
\newblock Minimal recursion semantics: An introduction.
\newblock \emph{Research on language and computation}, 3:\penalty0 281--332,
  2005.

\bibitem[Damonte et~al.(2019)Damonte, Goel, and Chung]{damonte.m.2019}
Marco Damonte, Rahul Goel, and Tagyoung Chung.
\newblock Practical semantic parsing for spoken language understanding.
\newblock In \emph{Proceedings of the 2019 Conference of the North American
  Chapter of the Association for Computational Linguistics: Human Language
  Technologies, Volume 2 (Industry Papers)}, pp.\  16--23, 2019.

\bibitem[Dong \& Lapata(2016)Dong and Lapata]{dong.l.2016}
Li~Dong and Mirella Lapata.
\newblock Language to logical form with neural attention.
\newblock In \emph{Proceedings of the 54th Annual Meeting of the Association
  for Computational Linguistics (Volume 1: Long Papers)}, pp.\  33--43, Berlin,
  Germany, 2016. Association for Computational Linguistics.
\newblock \doi{10.18653/v1/P16-1004}.
\newblock URL \url{https://aclanthology.org/P16-1004}.

\bibitem[Dwivedi(2013)]{dwivedi.v.2013}
Veena~D Dwivedi.
\newblock Interpreting quantifier scope ambiguity: Evidence of heuristic first,
  algorithmic second processing.
\newblock \emph{PloS one}, 8\penalty0 (11):\penalty0 e81461, 2013.

\bibitem[Filik et~al.(2004)Filik, Paterson, and Liversedge]{filik.r.2004}
Ruth Filik, Kevin~B Paterson, and Simon~P Liversedge.
\newblock Processing doubly quantified sentences: Evidence from eye movements.
\newblock \emph{Psychonomic Bulletin \& Review}, 11\penalty0 (5):\penalty0
  953--959, 2004.

\bibitem[Fodor \& Pylyshyn(1988)Fodor and Pylyshyn]{fodor.j.1988}
Jerry~A Fodor and Zenon~W Pylyshyn.
\newblock Connectionism and cognitive architecture: A critical analysis.
\newblock \emph{Cognition}, 28\penalty0 (1-2):\penalty0 3--71, 1988.

\bibitem[Futeral et~al.(2022)Futeral, Schmid, Laptev, Sagot, and
  Bawden]{futeral.m.2022}
Matthieu Futeral, Cordelia Schmid, Ivan Laptev, Beno{\^\i}t Sagot, and Rachel
  Bawden.
\newblock Tackling ambiguity with images: Improved multimodal machine
  translation and contrastive evaluation.
\newblock \emph{arXiv preprint arXiv:2212.10140}, 2022.

\bibitem[Kate et~al.(2005)Kate, Wong, Mooney, et~al.]{kate.r.2005}
Rohit~J Kate, Yuk~Wah Wong, Raymond~J Mooney, et~al.
\newblock Learning to transform natural to formal languages.
\newblock In \emph{AAAI}, volume~5, pp.\  1062--1068, 2005.

\bibitem[Kim \& Linzen(2020)Kim and Linzen]{kim.n.2020}
Najoung Kim and Tal Linzen.
\newblock {COGS}: A compositional generalization challenge based on semantic
  interpretation.
\newblock In \emph{Proceedings of the 2020 Conference on Empirical Methods in
  Natural Language Processing (EMNLP)}, pp.\  9087--9105, Online, 2020.
  Association for Computational Linguistics.
\newblock \doi{10.18653/v1/2020.emnlp-main.731}.
\newblock URL \url{https://aclanthology.org/2020.emnlp-main.731}.

\bibitem[Kurtzman \& MacDonald(1993)Kurtzman and MacDonald]{kurtzman.h.1993}
Howard~S Kurtzman and Maryellen~C MacDonald.
\newblock Resolution of quantifier scope ambiguities.
\newblock \emph{Cognition}, 48\penalty0 (3):\penalty0 243--279, 1993.

\bibitem[Lackner \& Garrett(1972)Lackner and Garrett]{lackner.j.1972}
James~R Lackner and Merrill~F Garrett.
\newblock Resolving ambiguity: Effects of biasing context in the unattended
  ear.
\newblock \emph{Cognition}, 1\penalty0 (4):\penalty0 359--372, 1972.

\bibitem[Lake \& Baroni(2018)Lake and Baroni]{lake.b.2018}
Brenden~M. Lake and Marco Baroni.
\newblock Generalization without systematicity: On the compositional skills of
  sequence-to-sequence recurrent networks.
\newblock In Jennifer~G. Dy and Andreas Krause (eds.), \emph{Proceedings of the
  35th International Conference on Machine Learning, {ICML} 2018,
  Stockholmsm{\"{a}}ssan, Stockholm, Sweden, July 10-15, 2018}, volume~80 of
  \emph{Proceedings of Machine Learning Research}, pp.\  2879--2888. {PMLR},
  2018.
\newblock URL \url{http://proceedings.mlr.press/v80/lake18a.html}.

\bibitem[Liu et~al.(2023)Liu, Wu, Michael, Suhr, West, Koller, Swayamdipta,
  Smith, and Choi]{liu.a.2023}
Alisa Liu, Zhaofeng Wu, Julian Michael, Alane Suhr, Peter West, Alexander
  Koller, Swabha Swayamdipta, Noah~A Smith, and Yejin Choi.
\newblock We're afraid language models aren't modeling ambiguity.
\newblock \emph{arXiv preprint arXiv:2304.14399}, 2023.

\bibitem[Mehrabi et~al.(2022)Mehrabi, Goyal, Verma, Dhamala, Kumar, Hu, Chang,
  Zemel, Galstyan, and Gupta]{mehrabi.n.2022}
Ninareh Mehrabi, Palash Goyal, Apurv Verma, Jwala Dhamala, Varun Kumar, Qian
  Hu, Kai-Wei Chang, Richard Zemel, Aram Galstyan, and Rahul Gupta.
\newblock Is the elephant flying? resolving ambiguities in text-to-image
  generative models.
\newblock \emph{arXiv preprint arXiv:2211.12503}, 2022.

\bibitem[Mialon et~al.(2023)Mialon, Dess{\`\i}, Lomeli, Nalmpantis, Pasunuru,
  Raileanu, Rozi{\`e}re, Schick, Dwivedi-Yu, Celikyilmaz,
  et~al.]{mialon.g.2023}
Gr{\'e}goire Mialon, Roberto Dess{\`\i}, Maria Lomeli, Christoforos Nalmpantis,
  Ram Pasunuru, Roberta Raileanu, Baptiste Rozi{\`e}re, Timo Schick, Jane
  Dwivedi-Yu, Asli Celikyilmaz, et~al.
\newblock Augmented language models: a survey.
\newblock \emph{arXiv preprint arXiv:2302.07842}, 2023.

\bibitem[Min et~al.(2020)Min, Michael, Hajishirzi, and Zettlemoyer]{min.s.2020}
Sewon Min, Julian Michael, Hannaneh Hajishirzi, and Luke Zettlemoyer.
\newblock Ambig{QA}: Answering ambiguous open-domain questions.
\newblock In \emph{Proceedings of the 2020 Conference on Empirical Methods in
  Natural Language Processing (EMNLP)}, pp.\  5783--5797, 2020.

\bibitem[Montague(1970)]{montague.r.1970}
Richard Montague.
\newblock English as a formal language.
\newblock \emph{Linguaggi nella Societa e nella Tecnica}, 1970.

\bibitem[Nie et~al.(2020)Nie, Zhou, and Bansal]{nie.y.2020}
Yixin Nie, Xiang Zhou, and Mohit Bansal.
\newblock What can we learn from collective human opinions on natural language
  inference data?
\newblock In \emph{Proceedings of the 2020 Conference on Empirical Methods in
  Natural Language Processing (EMNLP)}. Association for Computational
  Linguistics, 2020.

\bibitem[Nijkamp et~al.(2022)Nijkamp, Pang, Hayashi, Tu, Wang, Zhou, Savarese,
  and Xiong]{nijkamp.n.2022}
Erik Nijkamp, Bo~Pang, Hiroaki Hayashi, Lifu Tu, Huan Wang, Yingbo Zhou, Silvio
  Savarese, and Caiming Xiong.
\newblock Codegen: An open large language model for code with multi-turn
  program synthesis.
\newblock \emph{arXiv preprint arXiv:2203.13474}, 2022.

\bibitem[Ouyang et~al.(2022)Ouyang, Wu, Jiang, Almeida, Wainwright, Mishkin,
  Zhang, Agarwal, Slama, Ray, et~al.]{ouyang.l.2022}
Long Ouyang, Jeffrey Wu, Xu~Jiang, Diogo Almeida, Carroll Wainwright, Pamela
  Mishkin, Chong Zhang, Sandhini Agarwal, Katarina Slama, Alex Ray, et~al.
\newblock Training language models to follow instructions with human feedback.
\newblock \emph{Advances in Neural Information Processing Systems},
  35:\penalty0 27730--27744, 2022.

\bibitem[Parisi et~al.(2022)Parisi, Zhao, and Fiedel]{parisi.a.2022}
Aaron Parisi, Yao Zhao, and Noah Fiedel.
\newblock Talm: Tool augmented language models.
\newblock \emph{arXiv preprint arXiv:2205.12255}, 2022.

\bibitem[Parsons(1990)]{parsons.t.1990}
Terence Parsons.
\newblock \emph{Events in the Semantics of English: A Study in Subatomic
  Semantics}.
\newblock MIT Press, 1990.

\bibitem[Pavlick \& Kwiatkowski(2019)Pavlick and Kwiatkowski]{pavlick.e.2019}
Ellie Pavlick and Tom Kwiatkowski.
\newblock Inherent disagreements in human textual inferences.
\newblock \emph{Transactions of the Association for Computational Linguistics},
  7:\penalty0 677--694, 2019.

\bibitem[Piantadosi et~al.(2012)Piantadosi, Tily, and
  Gibson]{piantadosi.s.2012}
Steven~T Piantadosi, Harry Tily, and Edward Gibson.
\newblock The communicative function of ambiguity in language.
\newblock \emph{Cognition}, 122\penalty0 (3):\penalty0 280--291, 2012.

\bibitem[Radford et~al.(2019)Radford, Wu, Child, Luan, Amodei, Sutskever,
  et~al.]{radford.a.2019}
Alec Radford, Jeffrey Wu, Rewon Child, David Luan, Dario Amodei, Ilya
  Sutskever, et~al.
\newblock Language models are unsupervised multitask learners, 2019.

\bibitem[Rasmussen \& Schuler(2020)Rasmussen and Schuler]{rasmussen.n.2020}
Nathan~Ellis Rasmussen and William Schuler.
\newblock A corpus of encyclopedia articles with logical forms.
\newblock In \emph{Proceedings of the 12th Conference on Language Resources and
  Evaluation (LREC 2020)}, 2020.

\bibitem[Rayner et~al.(1983)Rayner, Carlson, and Frazier]{rayner.k.1983}
Keith Rayner, Marcia Carlson, and Lyn Frazier.
\newblock The interaction of syntax and semantics during sentence processing:
  Eye movements in the analysis of semantically biased sentences.
\newblock \emph{Journal of verbal learning and verbal behavior}, 22\penalty0
  (3):\penalty0 358--374, 1983.

\bibitem[Roy et~al.(2022)Roy, Thomson, Chen, Shin, Pauls, Eisner, and
  Van~Durme]{roy.s.2022}
Subhro Roy, Sam Thomson, Tongfei Chen, Richard Shin, Adam Pauls, Jason Eisner,
  and Benjamin Van~Durme.
\newblock Bench{CLAMP}: A benchmark for evaluating language models on semantic
  parsing.
\newblock \emph{arXiv preprint arXiv:2206.10668}, 2022.

\bibitem[Rudinger \& Van~Durme(2014)Rudinger and Van~Durme]{rudinger.r.2014}
Rachel Rudinger and Benjamin Van~Durme.
\newblock Is the stanford dependency representation semantic?
\newblock In \emph{Proceedings of the Second Workshop on EVENTS: Definition,
  Detection, Coreference, and Representation}, pp.\  54--58, 2014.

\bibitem[Sakaguchi \& Van~Durme(2018)Sakaguchi and Van~Durme]{sakaguchi.k.2018}
Keisuke Sakaguchi and Benjamin Van~Durme.
\newblock Efficient online scalar annotation with bounded support.
\newblock In \emph{Proceedings of the 56th Annual Meeting of the Association
  for Computational Linguistics (Volume 1: Long Papers)}, pp.\  208--218,
  Melbourne, Australia, July 2018. Association for Computational Linguistics.
\newblock \doi{10.18653/v1/P18-1020}.
\newblock URL \url{https://aclanthology.org/P18-1020}.

\bibitem[Schick et~al.(2023)Schick, Dwivedi-Yu, Dess{\`\i}, Raileanu, Lomeli,
  Zettlemoyer, Cancedda, and Scialom]{schick.t.2023}
Timo Schick, Jane Dwivedi-Yu, Roberto Dess{\`\i}, Roberta Raileanu, Maria
  Lomeli, Luke Zettlemoyer, Nicola Cancedda, and Thomas Scialom.
\newblock Toolformer: Language models can teach themselves to use tools.
\newblock \emph{arXiv preprint arXiv:2302.04761}, 2023.

\bibitem[Schutze(1995)]{schutze.h.1995}
Hinrich Schutze.
\newblock \emph{Ambiguity in language learning: {C}omputational and cognitive
  models}.
\newblock Stanford University, 1995.

\bibitem[{Semantic Machines} et~al.(2020){Semantic Machines}, Andreas, Bufe,
  Burkett, Chen, Clausman, Crawford, Crim, DeLoach, Dorner, Eisner, Fang, Guo,
  Hall, Hayes, Hill, Ho, Iwaszuk, Jha, Klein, Krishnamurthy, Lanman, Liang,
  Lin, Lintsbakh, McGovern, Nisnevich, Pauls, Petters, Read, Roth, Roy, Rusak,
  Short, Slomin, Snyder, Striplin, Su, Tellman, Thomson, Vorobev, Witoszko,
  Wolfe, Wray, Zhang, and Zotov]{semanticmachines2020}
{Semantic Machines}, Jacob Andreas, John Bufe, David Burkett, Charles Chen,
  Josh Clausman, Jean Crawford, Kate Crim, Jordan DeLoach, Leah Dorner, Jason
  Eisner, Hao Fang, Alan Guo, David Hall, Kristin Hayes, Kellie Hill, Diana Ho,
  Wendy Iwaszuk, Smriti Jha, Dan Klein, Jayant Krishnamurthy, Theo Lanman,
  Percy Liang, Christopher~H. Lin, Ilya Lintsbakh, Andy McGovern, Aleksandr
  Nisnevich, Adam Pauls, Dmitrij Petters, Brent Read, Dan Roth, Subhro Roy,
  Jesse Rusak, Beth Short, Div Slomin, Ben Snyder, Stephon Striplin, Yu~Su,
  Zachary Tellman, Sam Thomson, Andrei Vorobev, Izabela Witoszko, Jason Wolfe,
  Abby Wray, Yuchen Zhang, and Alexander Zotov.
\newblock Task-oriented dialogue as dataflow synthesis.
\newblock \emph{Transactions of the Association for Computational Linguistics},
  8:\penalty0 556--571, September 2020.
\newblock URL \url{https://doi.org/10.1162/tacl_a_00333}.

\bibitem[Shin \& Van~Durme(2022)Shin and Van~Durme]{shin.r.2022}
Richard Shin and Benjamin Van~Durme.
\newblock Few-shot semantic parsing with language models trained on code.
\newblock In \emph{Proceedings of the 2022 Conference of the North American
  Chapter of the Association for Computational Linguistics: Human Language
  Technologies}, pp.\  5417--5425, 2022.

\bibitem[Shin et~al.(2021)Shin, Lin, Thomson, Chen~Jr, Roy, Platanios, Pauls,
  Klein, Eisner, and Van~Durme]{shin.r.2021}
Richard Shin, Christopher Lin, Sam Thomson, Charles Chen~Jr, Subhro Roy,
  Emmanouil~Antonios Platanios, Adam Pauls, Dan Klein, Jason Eisner, and
  Benjamin Van~Durme.
\newblock Constrained language models yield few-shot semantic parsers.
\newblock In \emph{Proceedings of the 2021 Conference on Empirical Methods in
  Natural Language Processing}, pp.\  7699--7715, 2021.

\bibitem[Steedman(2011)]{steedman.m.2011}
Mark Steedman.
\newblock \emph{Taking scope: The natural semantics of quantifiers}.
\newblock Mit Press, 2011.

\bibitem[Stengel-Eskin \& Van~Durme(2022)Stengel-Eskin and
  Van~Durme]{stengel-eskin.e.2023calibration}
Elias Stengel-Eskin and Benjamin Van~Durme.
\newblock Calibrated interpretation: Confidence estimation in semantic parsing.
\newblock \emph{arXiv preprint arXiv:2211.07443}, 2022.

\bibitem[Stengel-Eskin \& Van~Durme(2023)Stengel-Eskin and
  Van~Durme]{stengel-eskin.e.2023didyoumean}
Elias Stengel-Eskin and Benjamin Van~Durme.
\newblock Did you mean...? {C}onfidence-based trade-offs in semantic parsing.
\newblock \emph{arXiv preprint arXiv:2303.16857}, 2023.

\bibitem[Stengel-Eskin et~al.(2023)Stengel-Eskin, Guallar-Blasco, Zhou, and
  Van~Durme]{stengel-eskin.e.2023}
Elias Stengel-Eskin, Jimena Guallar-Blasco, Yi~Zhou, and Benjamin Van~Durme.
\newblock Why did the chicken cross the road? {R}ephrasing and analyzing
  ambiguous questions in vqa.
\newblock \emph{Proceedings of the 61st Annual Meeting of the Association for
  Computational Linguistics}, 2023.

\bibitem[Tellex et~al.(2011)Tellex, Kollar, Dickerson, Walter, Banerjee,
  Teller, and Roy]{tellex.s.2011}
Stefanie Tellex, Thomas Kollar, Steven Dickerson, Matthew Walter, Ashis
  Banerjee, Seth Teller, and Nicholas Roy.
\newblock Understanding natural language commands for robotic navigation and
  mobile manipulation.
\newblock In \emph{Proceedings of the AAAI Conference on Artificial
  Intelligence}, volume~25, pp.\  1507--1514, 2011.

\bibitem[Tellex et~al.(2020)Tellex, Gopalan, Kress-Gazit, and
  Matuszek]{tellex.s.2020}
Stefanie Tellex, Nakul Gopalan, Hadas Kress-Gazit, and Cynthia Matuszek.
\newblock Robots that use language.
\newblock \emph{Annual Review of Control, Robotics, and Autonomous Systems},
  3:\penalty0 25--55, 2020.

\bibitem[Touvron et~al.(2023)Touvron, Lavril, Izacard, Martinet, Lachaux,
  Lacroix, Rozi{\`e}re, Goyal, Hambro, Azhar, et~al.]{touvron.h.2023}
Hugo Touvron, Thibaut Lavril, Gautier Izacard, Xavier Martinet, Marie-Anne
  Lachaux, Timoth{\'e}e Lacroix, Baptiste Rozi{\`e}re, Naman Goyal, Eric
  Hambro, Faisal Azhar, et~al.
\newblock Llama: Open and efficient foundation language models.
\newblock \emph{arXiv preprint arXiv:2302.13971}, 2023.

\bibitem[Vashishtha et~al.(2019)Vashishtha, Van~Durme, and
  White]{vashishtha.s.2019}
Siddharth Vashishtha, Benjamin Van~Durme, and Aaron~Steven White.
\newblock Fine-grained temporal relation extraction.
\newblock In \emph{Proceedings of the 57th Annual Meeting of the Association
  for Computational Linguistics}, pp.\  2906--2919, Florence, Italy, 2019.
  Association for Computational Linguistics.
\newblock \doi{10.18653/v1/P19-1280}.
\newblock URL \url{https://aclanthology.org/P19-1280}.

\bibitem[Wei et~al.(2022)Wei, Bosma, Zhao, Guu, Yu, Lester, Du, Dai, and
  Le]{wei.j.2022b}
Jason Wei, Maarten Bosma, Vincent Zhao, Kelvin Guu, Adams~Wei Yu, Brian Lester,
  Nan Du, Andrew~M Dai, and Quoc~V Le.
\newblock Finetuned language models are zero-shot learners.
\newblock In \emph{International Conference on Learning Representations}, 2022.

\bibitem[Winograd(1972)]{winograd.t.1972}
Terry Winograd.
\newblock Understanding natural language.
\newblock \emph{Cognitive psychology}, 3\penalty0 (1):\penalty0 1--191, 1972.

\bibitem[Wittgenstein(1921)]{wittgenstein.l.1921}
Ludwig Wittgenstein.
\newblock \emph{Tractatus Logico-Philosophicus}.
\newblock Annalen der Naturphilosophie, 1921.

\bibitem[Wu et~al.(2023)Wu, Manning, and Potts]{wu.z.2023}
Zhengxuan Wu, Christopher~D Manning, and Christopher Potts.
\newblock Re{COGS}: How incidental details of a logical form overshadow an
  evaluation of semantic interpretation.
\newblock \emph{arXiv preprint arXiv:2303.13716}, 2023.

\bibitem[Yu et~al.(2018)Yu, Zhang, Yang, Yasunaga, Wang, Li, Ma, Li, Yao,
  Roman, et~al.]{yu.t.2018}
Tao Yu, Rui Zhang, Kai Yang, Michihiro Yasunaga, Dongxu Wang, Zifan Li, James
  Ma, Irene Li, Qingning Yao, Shanelle Roman, et~al.
\newblock Spider: A large-scale human-labeled dataset for complex and
  cross-domain semantic parsing and text-to-sql task.
\newblock In \emph{Proceedings of the 2018 Conference on Empirical Methods in
  Natural Language Processing}, pp.\  3911--3921, 2018.

\bibitem[Yuan et~al.(2023)Yuan, Malaviya, and Yatskar]{yuan.y.2023}
Yuewei Yuan, Chaitanya Malaviya, and Mark Yatskar.
\newblock Ambi{C}oref: Evaluating human and model sensitivity to ambiguous
  coreference.
\newblock In \emph{Findings of the Association for Computational Linguistics:
  EACL 2023}, pp.\  993--1000, 2023.

\bibitem[Zelle \& Mooney(1996)Zelle and Mooney]{zelle.j.1996}
John~M Zelle and Raymond~J Mooney.
\newblock Learning to parse database queries using inductive logic programming.
\newblock In \emph{Proceedings of the national conference on artificial
  intelligence}, pp.\  1050--1055, 1996.

\bibitem[Zhang et~al.(2019)Zhang, Ma, Duh, and Van~Durme]{zhang.s.2019b}
Sheng Zhang, Xutai Ma, Kevin Duh, and Benjamin Van~Durme.
\newblock Broad-coverage semantic parsing as transduction.
\newblock In \emph{Proceedings of the 2019 Conference on Empirical Methods in
  Natural Language Processing and the 9th International Joint Conference on
  Natural Language Processing (EMNLP-IJCNLP)}, pp.\  3786--3798, Hong Kong,
  China, 2019. Association for Computational Linguistics.
\newblock \doi{10.18653/v1/D19-1392}.
\newblock URL \url{https://aclanthology.org/D19-1392}.

\bibitem[Zipf(1949)]{zipf.g.1949}
George~Kingsley Zipf.
\newblock \emph{Human Behaviour and the Principle of Least Effort}.
\newblock Addison-Wesley Press, 1949.

\end{thebibliography}
\bibliographystyle{iclr2024_conference}

\appendix
\section{Appendix}

\subsection{Dataset Construction} \label{append:data}
We take a neo-Davidsonian event semantics approach \citep{parsons.t.1990} to our logical forms (LFs), expressing our logical forms in quantified first-order logic (FOL). 
Events are represented as variables, with event-type predicates applied to them. 
For example, the statement \emph{a woman walks} would be represented as $\exists x . \exists e . woman(x) \land walk(e) \land agent(e,x)$, allowing for an arbitrary number of semantic roles; the semantic roles covered in our dataset are \emph{agent}, \emph{patient}, and \emph{instrument}.
Generic noun phrases like \emph{``a dog''} are existentially quantified: $\exists x . dog(x)$.
Proper nouns are assumed to have a single referrent, and are not quantified, e.g. \emph{Mary walks} $\rightarrow \exists e . walks(e) \land agent(e, Mary)$.

\subsubsection{Lexical Items} \label{append:lex} For PP ambiguity, we pair visual verbs (e.g. \emph{see, observe, spot, etc.}) with visual instruments (e.g. \emph{telescope, binoculars, etc.}) and tactile verbs (e.g. \emph{grab, pick up, etc.}) with things that can be worn/possessed and used for manipulation (e.g. \emph{gloves, ovenmitts, tongs, etc.}). For scope and reverse scope, we use common nouns and visual and tactile verbs. For pronoun coreference, the lexical items used here are gendered names (e.g \emph{Mary, John, etc.}) and gendered nouns (e.g. \emph{woman, man, boy, girl}). Conjunction examples use intransitive verbs. 

\subsubsection{Existential quantification} For uses of \emph{``the''}, we differ from \citet{kim.n.2020}, who use the $\iota$ notation for definite articles to denote a uniqueness clause. 
This is implemented as a existential quantifier at the widest scope, which can be ignored in all cases except scope ambiguity, where we only have indefinitely-quantified NPs. 
Similarly, \citet{artzi.y.2015} introduce Skolem terms \citep{steedman.m.2011} for definite NPs, which are also globally scoped.
Thus, we do not differentiate between definite and indefinite NPs in \textsc{AmP}. This has no impact on ambiguity.

\subsubsection{Extending to new templates} The framework we release allows for the addition of new templates and lexical items. 
To add a new template, the user specifies a surface-form template and an LF template, and provides the set of lexical items that can be used to fill slots in the templates. 
The framework enumerates all possible combinations of lexical items which respect the template constraints and produces paired inputs and LFs. 

\subsection{Zero-shot Prompts} \label{append:prompt}

\paragraph{PP Attachment} For PP attachment, we show an example the main verb being used transitively, the instrumental use of ``with'', and the accompaniment use of ``with''. 
\begin{lstlisting}
Let's translate what a human user says into what a computer might say.


Human: Galileo saw Mary
Computer: exists a . saw(a) AND agent(a, Galileo) AND patient(a, Mary)

Human: Mary with the camera
Computer: exists x . exists a . camera(x) AND have(a) AND agent(a, Mary) AND patient(a, x)

Human: Galileo saw with the camera
Computer: exists x . exists a . camera(x) AND saw(a) AND agent(a, Galileo) AND instrument(a, x)

Human: Galileo saw Mary with the camera
Computer:
\end{lstlisting}

\paragraph{Conjunctions} For conjunction ambiguities, we include an example of double conjunction (e.g. \emph{and ... and}) and double disjunction (e.g. \emph{or ... or}). The bracketing can vary. 
\begin{lstlisting}
Let's translate what a human user says into what a computer might say.


Human: the bird left and walked and ate
Computer: exists x . exists a . exists e . exists i . bird(x) AND ( left(a) AND agent(a, x) AND walked(e) AND agent(e, x) ) AND ate(i) AND agent(i, x)

Human: the bird left or walked or ate
Computer: exists x . exists a . exists e . exists i . bird(x) AND ( left(a) AND agent(a, x) ) OR ( ( walked(e) AND agent(e, x) ) OR ( ate(i) AND agent(i, x) ) )

Human: the bird left or walked and ate
Computer:
\end{lstlisting}

\paragraph{Bound pronouns} For pronoun coreference, we show the transitive verb and each possible subject with the embedded verb separately. We also include an example of a subject with two verbs so that the model sees how to compose two verbs in the same sentence. 

\begin{lstlisting}
Let's translate what a human user says into what a computer might say.


Human: the woman saw Marie
Computer: exists x . exists a . woman(x) AND saw(a) AND agent(a, x) AND patient(a, Marie)

Human: Marie smiled
Computer: exists a . smiled(a) AND agent(a, Marie)

Human: the woman frowned and smiled
Computer: exists x . exists a . exists e . woman(x) AND frowned(a) AND agent(a, x) AND smiled(e) AND agent(e, x)

Human: the woman smiled
Computer: exists x . exists a . woman(x) AND smiled(a) AND agent(a, x)

Human: the woman saw Marie and she smiled
Computer:    
\end{lstlisting}

\paragraph{Scope} Scope ambiguity prompts include an example of the verb being used transitively, as well an example of universal quantification. 
\begin{lstlisting}
Let's translate what a human user says into what a computer might say.


Human: a bird held a sweater
Computer: exists x . exists y . exists a . bird(x) AND sweater(y) AND held(a) AND agent(a, x) AND patient(a, y)

Human: each bird
Computer: forall x . bird(x)

Human: each bird held a sweater
Computer:
\end{lstlisting}

\paragraph{Inverse scope} Inverse scope prompts include the same information as scope ambiguities but with reversed arguments. 
\begin{lstlisting}
Let's translate what a human user says into what a computer might say.


Human: a dog spotted a hat
Computer: exists x . exists y . exists a . dog(x) AND hat(y) AND spotted(a) AND agent(a, x) AND patient(a, y)

Human: each hat
Computer: forall x . hat(x)

Human: a dog spotted each hat
Computer: 
\end{lstlisting}

\subsection{Constrained decoding} \label{append:decoding}
For locally-run models, we use grammar-constrained decoding \citep{shin.r.2021, shin.r.2022, roy.s.2022} to ensure that the model produces syntactically-correct formulae. 
During decoding, we use the BenchCLAMP framework \citep{roy.s.2022} to restrict the model's output vocabulary according to a context-free grammar, such that the model can only produce strings accepted by the grammar.\footnote{We release our FOL and Lisp grammars.}
This allows us to separate the model's semantic performance from its syntactic abilities. 
We decode with beam search, using a beam of 5. 

\subsection{Zero-shot Parsing: Qualitative Analysis}
\rebut{
\cref{sec:zeroshot_res} shows that models typically perform poorly on zero-shot parsing. 
Given that we use constrained decoding on all open-source models, the errors they make cannot be syntactic in nature, i.e. their outputs are guaranteed to be well-formed FOL expressions.
This raises the question of what kinds of errors models are making. 
Here, we qualitatively analyze model errors. 
For each ambiguity type, we sample 10 incorrect examples from the Codegen-16B model and classify the errors the model makes. 
\begin{itemize}
\item \textbf{PP attachment:} There are two classes of errors. 9/10 examples have a missing predicate, e.g. \mytt{exists v0 . exists v1 . agent(v1, Adele) AND have(v1) AND instrument(v1, v0) AND spied(v1) AND telescope(v0)
} is missing a \mytt{patient(v1, Sherlock)} predicate for the sentence ``Adele spied Sherlock with a telescope''. The remaining example had incorrect variable usage. 
\item \textbf{Scope:} 9/10 incorrect examples had a $\exists$ in place of the $\forall$ quantifier (i.e. the right number of quantifiers but no $\forall$ quantifier). 1/10 was missing a $\exists$ quantifier. 
\item \textbf{Inverse Scope:} 10/10 examples had a $\exists$ in place of the $\forall$ quantifier.
\item \textbf{Bound: } 8/10 examples used the same variable for 2 verbs. For example, in \mytt{exists v0 . agent(v0, Katherine) AND frowned(v0) AND observed(v0) AND patient(v0, Mary)}, the event variable \mytt{v0} is used for \mytt{frowned} and \mytt{observed} when there should be an additional event variable \mytt{v1}. 1/10 examples had other incorrect variable use, and another had a different missing predicate. 
\item \textbf{Conjunctions: } 8/10 examples had the wrong connectives, i.e. \mytt{OR, OR, AND} instead of \mytt{OR, AND, AND}. 2/10 had bad scoping, where the correct predicates and connectives were produced but were grouped incorrectly. 
\end{itemize} }

\subsection{Annotation task} \label{append:annotation}
To ensure the validity of our results, we conducted a pilot paraphrasing task, where we asked Mechanical Turk annotators to verify that they were native English speakers and paraphrase a short passage. 
In the task, annotators from a trusted list were first asked if they were native English speakers. 
To verify the results, annotators were additionally asked to paraphrase a short fable (The North Wind and the Sun) in 3 sentences.
The annotation interface precluded copying, preventing annotators from using external resources. 
Annotators were paid $\$0.50$ for the pilot task, corresponding to an hourly payment of $\sim\$14.00$
Each annotator's summary was then manually checked to verify fluency and adequacy; all annotators passed the quality check. 

\begin{figure}[h]
    \centering
    \includegraphics[width=0.75\textwidth]{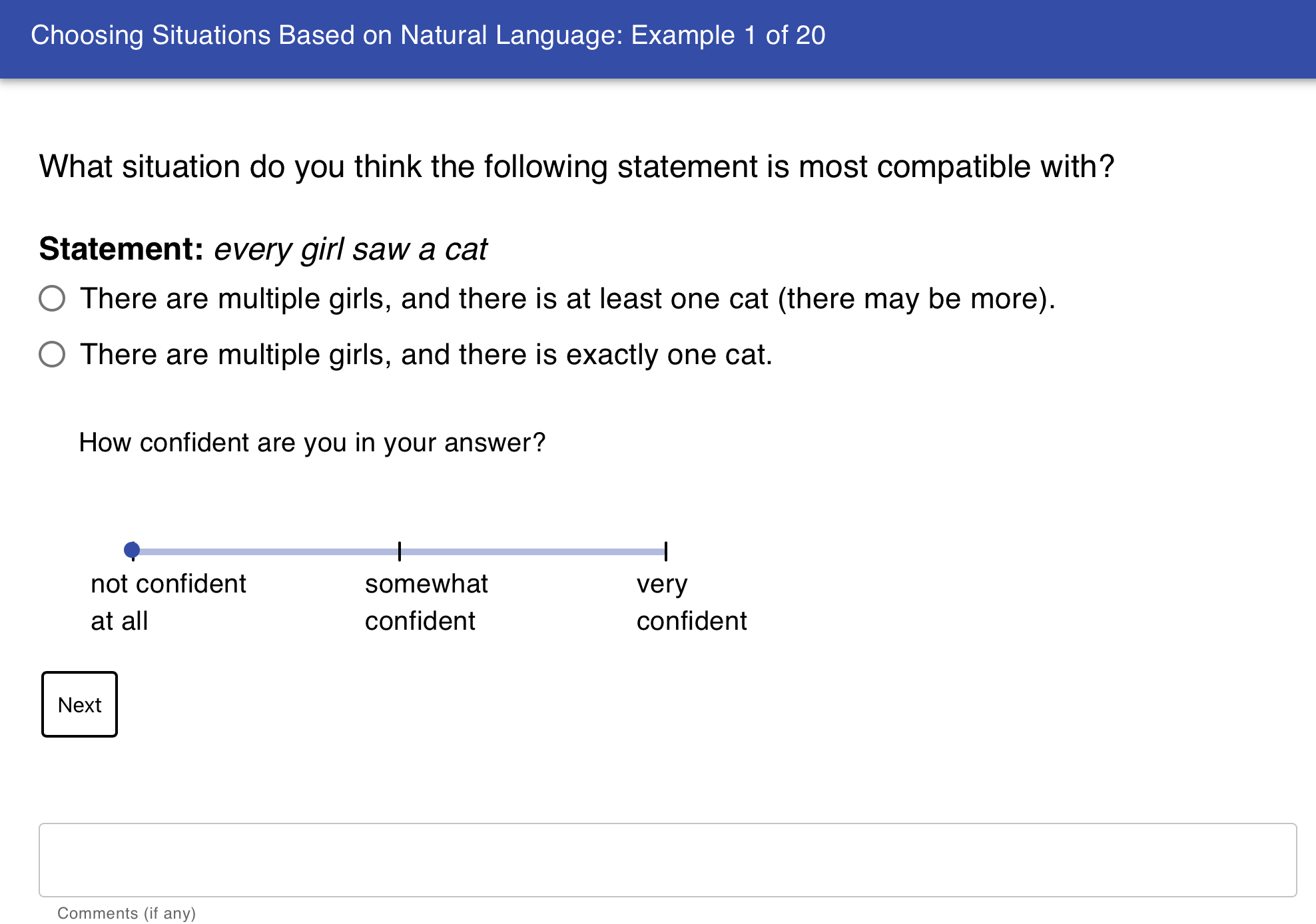}
    \caption{Annotation interface for human evaluation.}
    \label{fig:interface}
\end{figure}

In the main HIT, annotators were again paid $\sim \$14.00$ per hour, and each annotator performed exactly 20 annotations in a sequence. Each sequence has 5 tuples of the 4 ambiguity types.
Annotators were shown a sliding scale with 3 ticks: \emph{not confident}, \emph{somewhat confident}, \emph{very confident}, and annotators can slide the indicator anywhere along the scale. 
The interface can be seen in \cref{fig:interface}.

To transform the annotators' confidence scores into probabilities, we first min-max normalize raw confidence scores, following past work using sliding bars \citep{vashishtha.s.2019}.
This accounts for the fact that different annotators may use the slider differently.
We then take the lowest confidence value to correspond to $p(LF_c) = 0.5$, where $LF_c$ is the LF corresponding to the chosen interpretation. 
Intuitively, if $p(LF_c)$ were less than 0.5, the annotator would have chosen the other LF. 
The highest confidence value corresponds to $p(LF_c) = 1.0$. 

Qualitatively, we find that visual verbs and nouns (e.g. \emph{saw-telescope}, \emph{observed-glasses}) are matched more to $LF_1$, where the PP is an instrument, while tactile verbs and nouns (e.g. \emph{held-gloves}, \emph{picked up-mittens}) yield a possessive interpretation. 

\paragraph{Limitations of the Annotation Task}
\rebut{To gather human preferences, we elicit choices between verbalized interpretations of each logical form. 
This is a different task from what the models are being tasked with, and is motivated by the fact that annotators are unlikely to know first-order logic. 
Even if they did, it is difficult to constrain annotators to produce exactly the kind of first-order logic statements that would match the reference. 
In this sense, most of the models we test have an advantage, as we use constrained decoding according to a grammar. 
Thus, the model cannot produce LFs that deviate from the syntax expected by AMP. }

\rebut{Our method for eliciting judgements differs from standard methods in psycholinguistics, which are typically based on reading times, eye tracking, or other more elaborate experimental methods \citep{lackner.j.1972, rayner.k.1983, filik.r.2004, dwivedi.v.2013}. 
It is more akin to the paraphrase verification method used by \citet{rayner.k.1983} to elicit interpretations. 
Note that these methods test for different things. 
While the former set of methods typically test for incremental and subconscious processes, our method and paraphrase verification test for conscious, non-incremental judgements. 
In the context of a comparison to transformer-based models which do not receive incremental input, the second paradigm is a more accurate fit.
Nevertheless, the uncertainty consciously expressed by our human annotators may differ, for example, from the uncertainty we would obtain via more direct measurements like reading times.}

\end{document}